\documentclass{article}
\usepackage{ijcai26}
\usepackage{times}
\usepackage{soul}
\usepackage{url}
\usepackage[hidelinks]{hyperref}
\usepackage[utf8]{inputenc}
\usepackage[small]{caption}
\usepackage{graphicx}
\usepackage{amsmath}
\usepackage{amsthm}
\usepackage{booktabs}
\usepackage{algorithm}
\usepackage{algorithmic}
\usepackage{float}
\usepackage{tikz}
\usepackage{pgfplots}
\pgfplotsset{compat=1.17}
\usetikzlibrary{arrows.meta, positioning, fit, shadows, calc}

\title{Reaching the Tail: Calibration Diversity Drives Conformal Coverage under Data Scarcity}

\author{
Donald Aadithiyan\\
\affiliations
Department of Computer Science and Engineering, University of Moratuwa, Moratuwa, Sri Lanka\\
\emails
donaldaadithiyanwork@gmail.com
}

\begin{document}

\maketitle

\begin{abstract}
Multi-horizon rare-event forecasting is hard under long macroeconomic
series' data constraints: labeled events are scarce, and standard
uncertainty quantification assumes an exchangeability that
autocorrelation violates. A controlled ablation shows an apparent
rare-event threshold for Adaptive Conformal Inference instead reflects
calibration-set size. Across 200 random calibration sets, support
width of the nonconformity-score distribution explains up to 85\% of
coverage variance versus 2\% for rare-event count; the same
\emph{ranking}, not the same magnitude, replicates across synthetic
conditions and five countries (five-country Spearman $\rho$ 0.45--0.66 vs. 0.02--0.23). A diversity-maximizing selector built on this is the only
strategy tested that improves long-horizon coverage
(67.8\%$\to$81.4\% at six months); Mondrian, shift-robust, and
extreme-value alternatives fail to close it. Mondrian even worsens
coverage under oracle labels. A compact proposition explains why:
coverage deficit reflects how closely the calibration set's upper
quantile reaches the test distribution's. Diversity is necessary, not
sufficient. Demonstrated on a two-stage U.S.\ recession-forecasting
framework with RegressorChain, whether six-month coverage reaches 90\%
under honest scoring remains open, a question this paper quantifies
rather than resolves.
\end{abstract}

\section{Introduction}

Rare-event forecasting over long historical records is a recurring
difficulty in applied machine learning. Labeled events are few, the
underlying series are autocorrelated rather than exchangeable, and
predictions are often needed at several horizons at once rather than one.
U.S.\ recession forecasting is a useful setting to study this difficulty,
since it combines all three properties. Recessions occur in
fewer than 15\% of months, the record rarely predates the mid to late 1960s, and
policymakers need probability estimates at multiple horizons rather than a
single point forecast, while still offering enough history, compared with
most other countries, to evaluate candidate methods. The methods
developed here are tested on the U.S.\ and five further countries, all
with long, well-curated statistical records; whether they extend to
jurisdictions with shorter or noisier records is not yet tested.

This paper studies three properties of this setting: class imbalance,
since recessions are rare events requiring loss functions and calibration
strategies built around this; inter-temporal output dependency, since a
6-month-ahead forecast is not independent of the 1-month-ahead forecast
for the same event; and uncertainty quantification under temporal
dependence, since standard conformal methods assume an exchangeability
that autocorrelated, shifting macroeconomic series violate.

Rather than resolve these three properties in general, we evaluate a practical framework combining inter-temporal conditioning and
horizon-specific calibration diagnostics. An apparent rare-event threshold in conformal coverage does not hold once
tested under controlled conditions: this paper identifies what actually
drives coverage, builds a calibration method around it, compares that
method against the standard imbalanced-conformal toolbox, and checks
whether the relationship holds beyond one series and country. The
diagnostic finding is this paper's main contribution; the selector,
theory, and testbed below support and test it, This paper contributes:

\begin{itemize}
\item A diversity-maximizing calibration selector, the only strategy
tested that meaningfully improves long-horizon coverage, alongside its
cost in wider intervals.
\item A fixed-size calibration ablation that falsifies an
apparent rare-event threshold, with a quantitative analysis
showing calibration-set diversity, not rare-event count, is what
predicts conformal coverage.
\item A comparison against the standard imbalanced-conformal toolbox:
Mondrian, a shift-robust controller, and extreme-value tail
fitting all fail to close the coverage gap, and Mondrian makes coverage
worse even under oracle labels. Only the selector closes it, across
synthetic conditions and five countries.
\item A compact proposition explaining why diversity helps but is not
sufficient on its own: coverage deficit reflects how closely the
calibration set's upper quantile reaches the test distribution's.
\item A multi-horizon recession-forecasting testbed built around a
RegressorChain that conditions each horizon's forecast on shorter ones,
and a decoupled two-stage architecture.
\end{itemize}

\section{Related Work}

Classical work used macroeconomic leading indicators for recession
prediction: yield curve spreads \cite{estrella1998predicting} and probit
models on credit and yield signals \cite{bluwstein2023credit}. Later
work applied tree-based ensembles \cite{dopke2017predicting,omolo2024using},
penalized logistic regression \cite{chung2023real}, and neural sequence
models \cite{chung2023inside}, generally without inter-temporal
conditioning or uncertainty quantification. Adaptive Conformal Inference (ACI) \cite{gibbs2021adaptive} and its
time-series extensions \cite{zaffran2022adaptive,angelopoulos2023conformal}
relax the exchangeability assumption that structural breaks violate
\cite{barber2023beyond,barber2025predictive}. Weighted conformal prediction
\cite{tibshirani2019conformal} addresses this shortfall via importance
reweighting under a known shift; our proposition
(Section~\ref{sec:calib-method}) restates it without assuming one.
Class-wise (Mondrian) conformal calibration
\cite{vovk2003mondrian,shi2024rc3p} partitions the calibration set by
regime for class-conditional coverage. Section~\ref{sec:aci} tests it as
a regression baseline; it degrades coverage from test-window imbalance,
not the approach itself. Extreme-value conformal methods
\cite{pasche2026extreme} fit tail distributions to calibration scores
for high-confidence intervals in high-impact domains (flooding,
finance). A comparable baseline in Section~\ref{sec:aci} does not close
the gap either, showing the deficit is about diversity, not any one
method. The RegressorChain here extends classifier chains
\cite{read2011classifier,read2019classifier} from multi-label
classification to regression, conditioning each horizon's prediction on
previously predicted, not ground-truth, outputs.

\section{Methodology}

\subsection{Point Prediction Engine}

\begin{figure}[t]
\centering
\includegraphics[width=0.9\linewidth]{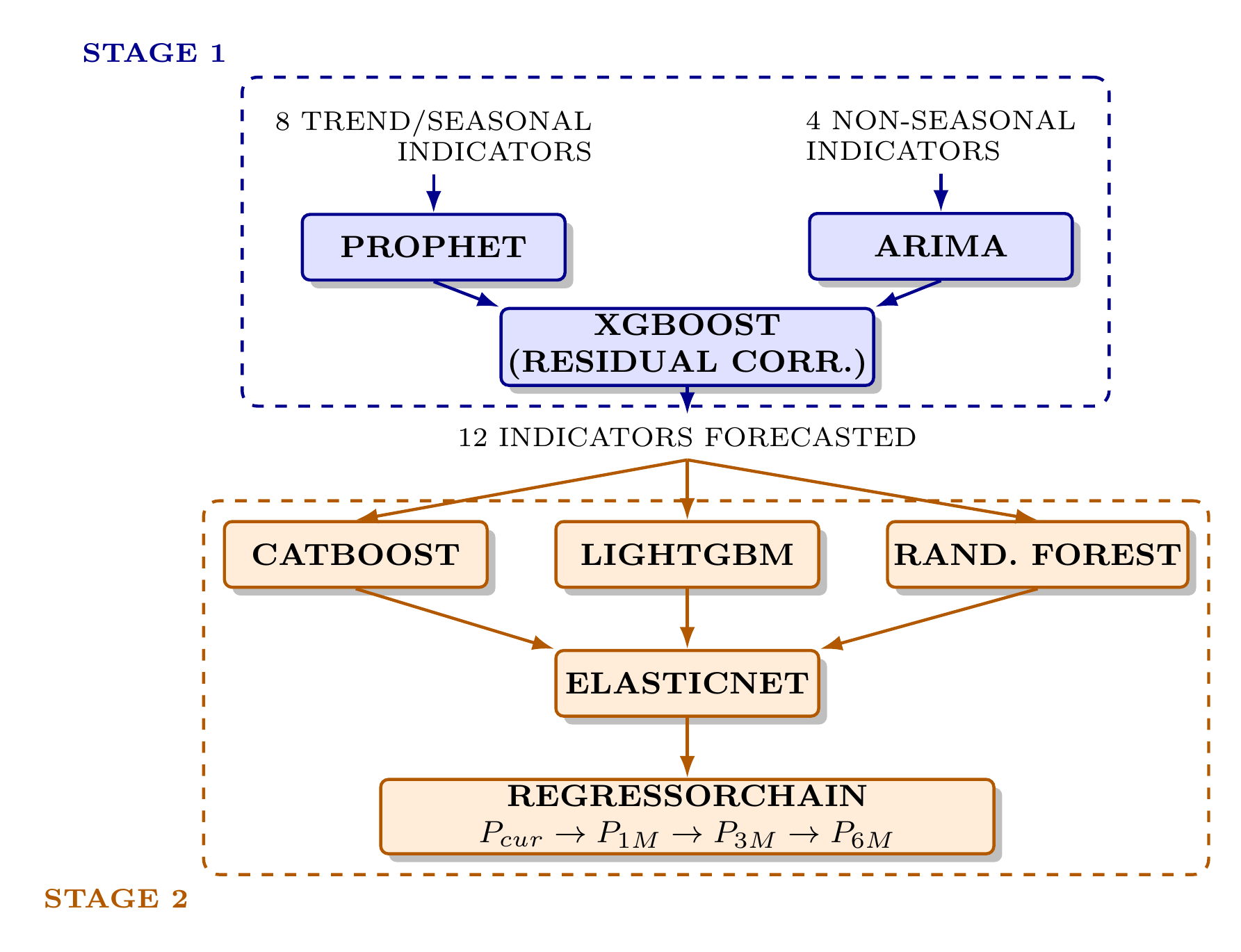}
\caption{Two-stage pipeline. Stage~1 forecasts indicators; Stage~2 maps
forecasts to horizon-conditioned probabilities.}
\label{fig:pipeline}
\end{figure}

The framework is a two-stage pipeline (Figure~\ref{fig:pipeline});
Appendix~\ref{app:components} separates which of its components the
central finding actually depends on. Stage~1
forecasts 12 macroeconomic indicators using hybrid Prophet and ARIMA
models with XGBoost residual correction. Stage~2 transforms these
forecasts into recession probabilities through a stacking ensemble with a
RegressorChain across horizons.

Data were obtained from the FRED API and cover 12 indicators: Treasury
rates at 1-, 3-, 6-, and 10-year maturities, CPI, PPI, Industrial
Production, Unemployment Rate, Share Price Index, GDP per Capita, OECD
CLI, and Consumer Sentiment. The recession target is FRED's smoothed
recession-probability series \cite{fred_recprousm156n}. Horizon-specific
targets are forward shifts of this series, at 1, 3, and 6 months, rather
than rolling averages, a distinction that matters for reading
Table~\ref{tab:baseline}. Data are split at January 2020, giving 635
training and 65 test observations while preserving temporal order; the
test period spans the COVID-19 recession and the 2022--2023 tightening
cycle. All preprocessing that could leak future information, including
cubic spline interpolation, Box-Cox transforms, STL decomposition, and
RFECV or Random Forest feature selection, is fit only on the pre-2020
partition. All lag, rolling, and autocorrelation features use a strictly
causal window, applying \texttt{shift(1)} before any rolling computation.

Prophet-XGBoost hybrids cover the 8 trend and seasonal indicators, and
ARIMA-XGBoost hybrids cover the remaining 4 (the 3-month Treasury rate,
PPI, Industrial Production, and Unemployment Rate), correcting
residuals via

$\hat{y}_t = \hat{y}_{base,t} + \hat{y}_{XGB,t}$. Stage~2 combines
CatBoost, LightGBM, and Random Forest through an ElasticNet meta-learner,
using 5-fold \texttt{TimeSeriesSplit} out-of-fold predictions, and then
chains the four horizons:
\begin{align*}
P_{cur} &= f_1(X), \quad P_{1M} = f_2(X, P_{cur}) \\
P_{3M} &= f_3(X, P_{cur}, P_{1M}), \quad P_{6M} = f_4(X, P_{cur}, P_{1M}, P_{3M})
\end{align*}
A logit transform combined with focal loss increases the weight given to
minority-class, or recession, periods during training.

\subsection{Diversity-Optimal Conformal Calibration}
\label{sec:calib-method}

Because macroeconomic series with structural breaks violate the
exchangeability assumption behind standard conformal prediction, this paper applies Adaptive Conformal Inference (ACI) \cite{gibbs2021adaptive}
(see Appendix~\ref{app:glossary} for a glossary of terms used below),
which
sets interval half-width from the empirical $(1{-}\alpha)$ quantile of
calibration-set nonconformity scores (the absolute forecast error
$|y_t - \hat{y}_t|$) and updates its miscoverage target
online. This section formalizes how that calibration set should be chosen
under rare-event scarcity. Section~\ref{sec:aci} tests this with data.

\textbf{Diversity-maximizing selection.} Given a fixed calibration budget
$N$, we select the $N$ months whose
pooled nonconformity scores maximize \emph{support width}
($p_{95}{-}p_5$), rather than the $N$ most recent months, the usual
trailing-window default, or the $N$ months with the most rare-event
episodes. The extreme-tail months provably maximize this quantity for a
fixed score set and count, so the selection is exact, not merely a greedy
heuristic. On this series, extreme-tail months are overwhelmingly rare-event months, so the selector's output overlaps heavily with a rare-event-maximizing
selector. This alone doesn't show that diversity and rare-event count are actually different things; that comes from the random-sampling analysis in
Section~\ref{sec:aci}.

\begin{figure}[H]
\centering
\includegraphics[width=\linewidth]{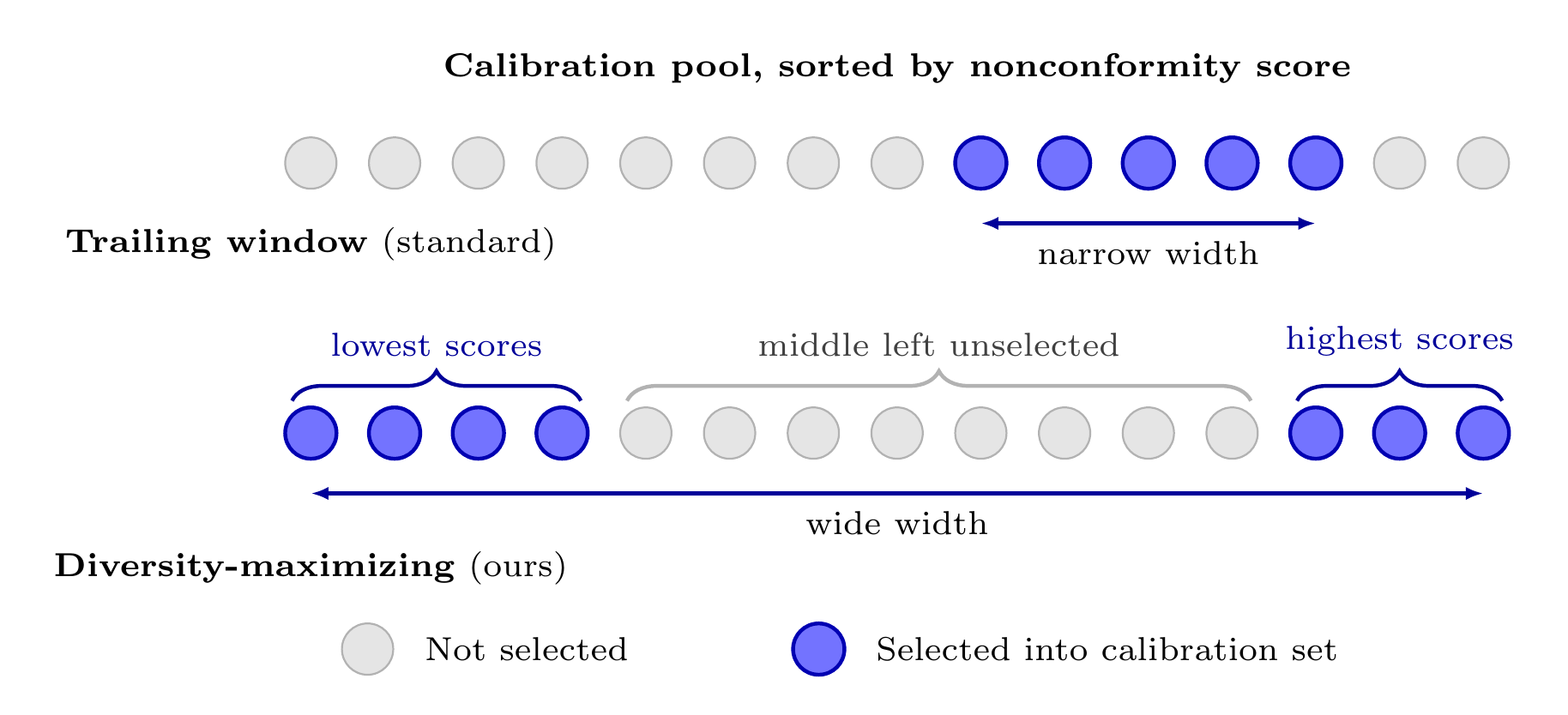}
\caption{The proposed selector versus the standard trailing window.
Selecting from both score tails, rather than the most recent months,
widens the calibration set's support, which is what drives the coverage
gain in Section~\ref{sec:aci} (an approximation, not a guarantee, of the
quantile reach that governs coverage).}
\label{fig:selection}
\end{figure}

\textbf{Why width, not composition.} Coverage fails when the calibration
set's tail does not reach as far as the test set's. Formally, let $\alpha$ (=0.10) be the target miscoverage level, $\hat{Q}_C(1{-}\alpha)$ the
calibration set's $(1{-}\alpha)$ quantile of nonconformity scores,
$Q_G(1{-}\alpha)$ the same quantile of the true test-time distribution,
and $G$ the function converting a score threshold into the coverage
probability it achieves. ACI's coverage deficit relative to nominal is
then 
\[
\Delta = G(Q_G(1{-}\alpha)) - G(\hat{Q}_C(1{-}\alpha)),
\]
monotone in the shortfall between $\hat{Q}_C$ and $Q_G$. Support width is a proxy for this quantile \emph{reach}, not the reach itself: a calibration set can look wide overall while still
missing the exact extreme the test period produces, $Q_G(1{-}\alpha)$
(Figure~\ref{fig:selection}). This is why the selector targets
extreme-tail months specifically rather than just variety in general,
and why it is necessary but not sufficient: it maximizes a proxy for
quantile reach, not the reach itself.

\section{Results and Discussion}

\subsection{The Drivers of Conformal Coverage under Data Scarcity}
\label{sec:aci}

\subsubsection{Composition, Not Size}

Sweeping a trailing window over five training-data fractions (20--40\%)
shows coverage jumping between 20\% and 25\%, resembling a rare-event
threshold. But window size and rare-event count change together here:
127 rows at 20\% versus 254 at 40\%, while rare-event months go from 1 to
16. Holding size fixed at $N{=}254$ and varying only the rare-event count
removes this confound (Table~\ref{tab:ablation}). Coverage then climbs gradually, and 1- and 3-month sit near ceiling
even with zero rare-event months, since small, stable short-horizon
errors let a narrow calibration set reach the test quantile; the
threshold's shape, not coverage, reflects window size.

\begin{table}[H]
\centering
\small
\begin{tabular}{lccccc}
\toprule
Rare mo. & Comp.\ \% & Cur. & 1M & 3M & 6M \\
\midrule
0  & 0.0 & 81.5 & 85.9 & 85.5 & 62.7 \\
1  & 0.4 & 83.1 & 85.9 & 85.5 & 62.7 \\
4  & 1.6 & 83.1 & 85.9 & 85.5 & 62.7 \\
8  & 3.1 & 84.6 & 85.9 & 85.5 & 66.1 \\
16 & 6.3 & 89.2 & 85.9 & 85.5 & 67.8 \\
\bottomrule
\end{tabular}
\caption{Coverage (\%) at fixed calibration size ($N{=}254$), varying the
rare-event months included; Comp.\ \% is their share of the window.}
\label{tab:ablation}
\end{table}

\subsubsection{Diversity Explains It}

Across 200 random $N{=}254$ calibration subsets, drawn without replacement
from the 635-month pool, support width ($p_{95}{-}p_5$) predicts coverage
far better than rare-event count. At 6 months support width ($R^2{=}0.85$)
beats rare-event count ($R^2{=}0.02$) by 50-fold, a gap that holds at 2 to
50 times for the 3- and 6-month horizons; Current and 1-month already
have little coverage variance to explain at this size. Because 200
draws share one fixed window, they aren't independent; $R^2$ is
descriptive, not a bound. Within diversity-matched subsets, the
remaining rare-event correlation collapses toward zero, pointing to
redundancy.

\textbf{The selector in practice.} Applying the diversity-maximizing
selector from Section~\ref{sec:calib-method} is the only strategy tested
that moves 6-month coverage at all, from 67.8\% to 81.4\%
(Figure~\ref{fig:methods}). It does so at the cost of much wider
intervals (10.6 to 32.6 points at six months), still falling short of
90\%, confirming diversity is necessary but not sufficient.

These numbers use in-sample training residuals. Recomputing with
out-of-fold scores, from a model refit at each point in time, gives a
similar gain: coverage rises from 84.75\% to 96.61\% at six months, a
gain of 11.9 points against the original 13.6. At this sample size,
though, neither figure is clearly distinguishable from the 90\% target:
both confidence intervals span it.

\subsubsection{Why Standard Alternatives Fail}
\begin{figure}[H]
\centering
\includegraphics[width=0.7\linewidth]{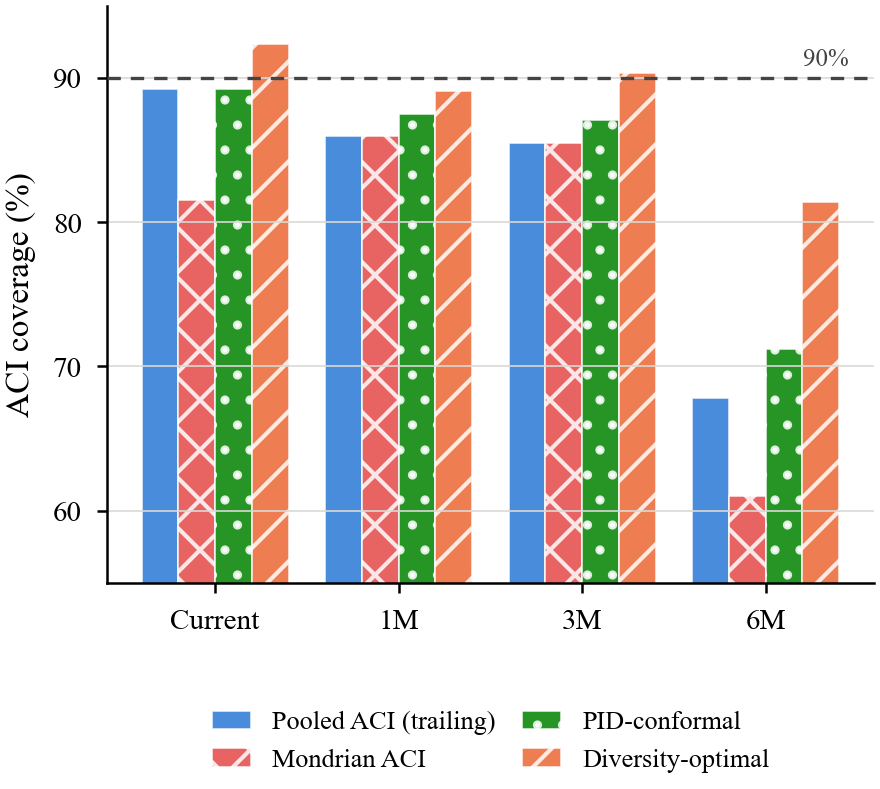}
\caption{ACI coverage by horizon: pooled (trailing), Mondrian, PID-conformal,
and diversity-optimal selection, against the 90\% nominal target. Mondrian
underperforms pooled ACI at every horizon; only diversity-optimal selection
moves 6-month coverage.}
\label{fig:methods}
\end{figure}

Mondrian, or class-conditional, ACI \cite{vovk2003mondrian} is the
standard fix for imbalanced conformal calibration, but it makes coverage
worse at every horizon, taking 6-month coverage from 67.8\% to 61.0\%,
even with oracle labels. The test window has 63 expansion
months and 2 rare-event months, so per-regime quantiles stop
borrowing the wide scores that pooled ACI relies on. Two more standard fixes fail as well: A shift-robust controller (PID-conformal) \cite{angelopoulos2023conformal}
and an extreme-value tail fit (a Generalized Pareto fit to scores above
the 80th percentile)
\cite{pasche2026extreme} also fail to close the gap
(Figure~\ref{fig:methods}). This points to a diversity problem, not a
weak algorithm; only the diversity-maximizing selector closes the gap.

\subsubsection{Generality}

Across 7 synthetic scenarios, nonconformity scores simulated directly
(rare/normal months drawn separately, varying frequency, magnitude,
and clustering, without the forecasting pipeline), Spearman's
$\rho$(diversity) ranges 0.42 to 0.68, always exceeding $\rho$(rare)
(0.02 to 0.45). The same ranking holds across five countries'
independent recession series (Table~\ref{tab:crosscountry}), a
separate dataset where diversity's own numbers run narrower
(0.45--0.66 vs.\ 0.02--0.23). Support width consistently out-predicts
rare-event count and the same redundancy check from above holds
throughout. The margin narrows at high frequency or magnitude (20\%,
6-fold), where rare-count and diversity converge near
$\rho{\approx}0.45$ in synthetic tests. The advantage is strongest in
the low-frequency, moderate-magnitude regime real recessions occupy
($\sim$8\% of months), not a general law across every rare-event
process.

\begin{table}[H]
\centering
\small
\begin{tabular}{lccc}
\toprule
Series & $\rho$(diversity) & $\rho$(rare) & Redund.$^*$ \\
\midrule
Euro area & 0.64 & 0.02 & 0.03 \\
UK        & 0.57 & 0.18 & 0.19 \\
Germany   & 0.45 & 0.10 & 0.10 \\
Japan     & 0.66 & 0.11 & 0.02 \\
Canada    & 0.65 & 0.23 & 0.09 \\
\bottomrule
\end{tabular}
\caption{Diversity vs.\ rare-event count as coverage predictors (Spearman
$\rho$), five independent countries' recession series. $^*$Within-diversity
$\rho$(rare, coverage); near zero confirms redundancy.}
\label{tab:crosscountry}
\end{table}

As predicted by the proposition in Section~\ref{sec:calib-method}, the
calibration quantile at 6 months falls well short of the test quantile
($\hat{Q}_C(0.90){=}3.01$ versus $Q_G(0.90){=}27.06$), predicting 57.6\%
coverage, 5 to 10 points below the 62.7--67.8\% observed across
composition and size strategies. The residual gap likely reflects
sampling noise and ACI's online updates. Coverage at 6 months remains
below 90\% in-sample for every strategy tested, an unresolved
limitation under structural breaks
\cite{barber2025predictive} that the out-of-fold check above
complicates.

\subsection{Testbed Evaluation: Point Prediction and Deployment}

Having established what governs conformal coverage under this rare-event
shift, this section validates the point-prediction testbed those results
were computed on.

\begin{table}[H]
\centering
\small
\begin{tabular}{lcccc}
\toprule
Model & Cur. & 1M & 3M & 6M \\
\midrule
Naive Mean            & 11.28 & 11.25 & 10.09 & 8.92$^{\dagger}$ \\
Probit (YC)            & 3.37  & 3.56  & 2.64  & 2.35 \\
XGB Indep.\ (tuned)     & 3.60  & 3.78  & 12.68 & 45.71$^{\ddagger}$ \\
MOR-XGB (tuned)         & 3.60  & 3.64  & 9.92  & 31.31 \\
MOR-XGB joint           & 3.13  & 3.84  & 10.08 & 28.59 \\
\textbf{RegressorChain} & \textbf{6.83} & \textbf{5.63} & \textbf{7.73} & \textbf{10.17} \\
\bottomrule
\end{tabular}
\caption{MAE (\%) on the 65-observation post-2020 test set. XGB baselines are
Optuna-tuned (40 trials, \texttt{TimeSeriesSplit}).}
\label{tab:baseline}
\end{table}

$^{\dagger}$Naive Mean's MAE falls with horizon since the 6M target
excludes the COVID-2020 spike, not better forecasting.

$^{\ddagger}$XGB Indep.'s 6-month failure is a regime-shift error
(cross-validation MAE 7.51, not misconfiguration): trained only on
pre-2020 data, it over-predicts risk through the 2022--2023 tightening.

Diebold-Mariano tests confirm the RegressorChain's advantage over tuned
XGB is significant at 3 and 6 months; a
block bootstrap confirms this at 6 months (disjoint 90\% CIs). Against
the probit, no horizon differs in point accuracy ($p{>}0.15$). Since ACI
and the diversity-maximizing selector are model-agnostic, we also apply
them directly to the probit's own residuals. The result favors neither
model: every strategy reaches coverage near or above 97\%, but only by
producing intervals 40 to 335 times wider than the target's actual
range. Coverage alone cannot show whether an interval is
well-calibrated or just too wide to be wrong.

\subsection{Limitations}

Four limitations qualify these results: this non-separation is specific
to this pool (Section~\ref{sec:calib-method}), and the selector's gain
costs wider intervals against deployment coverage; six-month coverage
remains unresolved under honest scoring, since the out-of-fold
estimate's confidence interval spans the 90\% target; the 65 post-2020
observations limit statistical power; and the mechanism replicates
across five countries (Section~\ref{sec:aci}), but the full pipeline is
U.S.-only.

\section{Conclusion}

\textbf{This work aims to improve rare-event uncertainty estimates for
practitioners in data-scarce settings, not to substitute for trained
economists' judgment or institutional policy processes.} The bottom
line: a diversity-maximizing selector outperforms the standard toolbox
(Mondrian, PID-conformal, extreme-value calibration), the only one that
improves long-horizon coverage here. The true driver is calibration-set
\emph{diversity}, not rare-event count. These findings suggest
calibration-set composition deserves the
same attention as calibration algorithms when designing conformal
predictors under scarcity. Practitioners should prioritize
diversity over raw size. Doing so brings the calibration set closer to
the tail, the rare, extreme values a forecast must anticipate. Whether
it reaches that tail at six months is, on the evidence so far, still an
open question.

\appendix

\section{Which Components Are Essential}
\label{app:components}

The paper combines several pieces: a two-stage forecasting pipeline, a
RegressorChain, ACI, and the diversity-maximizing selector. This
appendix separates what each piece is actually needed for.

The central finding, that calibration-set diversity predicts coverage
better than rare-event count, does not depend on the forecasting
pipeline or the RegressorChain at all. The synthetic-conditions check
(Section~\ref{sec:aci}, Generality) demonstrates this directly: scores
there are simulated, with no forecasting model in the loop, and the
same diversity-over-rare-count pattern holds. The probit comparison
(Section~\ref{sec:aci}, Testbed Evaluation) is a second, independent
check of the same point: applying ACI and the selector to a
completely different point-prediction model still produces the same
qualitative behaviour, even though that specific test window turned
out to be uninformative. Together, these show the diversity mechanism
is a property of ACI and the calibration set, not an artefact of any
one forecasting architecture.

The forecasting pipeline and RegressorChain matter for a separate
claim: point-prediction quality (Table~\ref{tab:baseline}). RegressorChain's
lower six-month MAE relative to the naive and independent-XGB baselines
is specific to that architecture and is not required for the
calibration-diversity finding to hold. In short: any competent
forecasting model, paired with ACI and the diversity-maximizing
selector, should reproduce the central result; the RegressorChain
pipeline is what makes the point predictions themselves good, a
different and separable contribution.

\section{Glossary of Terms}
\label{app:glossary}

Brief, plain-language definitions for terms used without full
introduction in the main text.

\textbf{RegressorChain.} A way of predicting several related targets,
here forecasts at different horizons, by feeding each earlier
prediction in as an input to the next, so the 3-month forecast can use
the current-month and 1-month forecasts rather than treating every
horizon independently.

\textbf{Nonconformity score.} How badly a model missed on one example,
here the absolute difference between prediction and outcome. Conformal
methods turn a distribution of past scores into a prediction interval
for future outcomes.

\textbf{Support width.} How far apart the smallest and largest
nonconformity scores in a calibration set are, measured as the
difference between their 95th and 5th percentiles so that a single
outlier cannot dominate. A wide support means the calibration set has
seen both calm and turbulent periods.

\textbf{Adaptive Conformal Inference (ACI).} A method that widens or
narrows its prediction intervals over time, based on whether recent
intervals actually contained the outcome, to target a chosen coverage
rate even as the data distribution shifts.

\textbf{Mondrian calibration.} Calibrating each class or regime
separately, for example recession months apart from expansion months,
instead of pooling all months together.

\textbf{PID-conformal.} A variant of ACI that adjusts interval width
using a control-theory feedback rule, reacting to both the current and
the accumulated coverage error, similar to a thermostat correcting for
persistent drift.

\textbf{Quantile reach.} Whether the calibration set's most extreme
scores are large enough to match the most extreme errors actually
encountered at test time. A calibration set can look diverse overall
and still fail to reach that far.

\textbf{Spearman $\rho$.} A correlation measure based on the rank order
of values rather than their raw size, capturing whether one quantity
tends to rise when another does, without assuming a straight-line
relationship.

\textbf{Focal loss.} A training loss that down-weights easy, common
examples so the model is pushed to pay proportionally more attention to
rare ones.

\textbf{Diebold-Mariano test.} A statistical test for whether one
forecaster's errors are genuinely smaller than another's, rather than
smaller by chance.

\textbf{Block bootstrap.} A resampling method that draws contiguous
blocks of consecutive observations, rather than single points, so that
autocorrelation within a time series is preserved in the resampled
data.

\bibliographystyle{named}
\bibliography{ijcai26}

@article{shi2024rc3p,
  author = "Yuanjie Shi and Subhankar Ghosh and Taha Belkhouja and Janardhan Rao Doppa and Yan Yan",
  title = "Conformal Prediction for Class-wise Coverage via Augmented Label Rank Calibration",
  journal = "Advances in Neural Information Processing Systems (NeurIPS)",
  year = "2024"
}

@article{read2011classifier,
  author = "Jesse Read and Bernhard Pfahringer and Geoff Holmes and Eibe Frank",
  title = "Classifier Chains for Multi-label Classification",
  journal = "Machine Learning",
  volume = "85",
  number = "3",
  pages = "333--359",
  year = "2011"
}

@article{read2019classifier,
  author = "Jesse Read and Bernhard Pfahringer and Geoff Holmes and Eibe Frank",
  title = "Classifier Chains: A Review and Perspectives",
  journal = "arXiv preprint arXiv:1912.13405",
  year = "2019"
}

@inproceedings{zaffran2022adaptive,
  author = "Margaux Zaffran and Olivier Feron and Yannig Goude and Julie Josse and Aymeric Dieuleveut",
  title = "Adaptive Conformal Predictions for Time Series",
  booktitle = "Proceedings of the 39th International Conference on Machine Learning (ICML)",
  series = "PMLR",
  volume = "162",
  year = "2022"
}

@inproceedings{angelopoulos2023conformal,
  author = "Anastasios N. Angelopoulos and Emmanuel J. Cand{\`e}s and Ryan J. Tibshirani",
  title = "Conformal {PID} Control for Time Series Prediction",
  booktitle = "Advances in Neural Information Processing Systems (NeurIPS)",
  year = "2023"
}

@article{estrella1998predicting,
  author = "Arturo Estrella and Frederic S. Mishkin",
  title = "Predicting {U.S.} Recessions: Financial Variables as Leading Indicators",
  journal = "Review of Economics and Statistics",
  volume = "80",
  number = "1",
  pages = "45--61",
  year = "1998"
}

@article{bluwstein2023credit,
  author = "Kristina Bluwstein and Marcus Buckmann and Andreas Joseph and Sujit Kapadia and {\"O}zg{\"u}r {\c{S}}im{\c{s}}ek",
  title = "Credit Growth, the Yield Curve, and Financial Crisis Prediction",
  journal = "Journal of International Economics",
  volume = "145",
  pages = "103773",
  year = "2023"
}

@article{dopke2017predicting,
  author = "J{\"o}rg D{\"o}pke and Ulrich Fritsche and Christian Pierdzioch",
  title = "Predicting Recessions with Boosted Regression Trees",
  journal = "International Journal of Forecasting",
  volume = "33",
  number = "4",
  pages = "745--759",
  year = "2017"
}

@article{omolo2024using,
  author = "Leakey Omolo and Nguyet Nguyen",
  title = "Using an Ensemble of Machine Learning Algorithms to Predict Economic Recession",
  journal = "Journal of Risk and Financial Management",
  volume = "17",
  number = "9",
  pages = "387",
  year = "2024"
}

@article{chung2023real,
  author = "Seulki Chung",
  title = "Real-time Prediction of the Great Recession and the {COVID-19} Recession",
  journal = "arXiv preprint arXiv:2310.08536",
  year = "2023"
}

@article{chung2023inside,
  author = "Seulki Chung",
  title = "Inside the Black Box: Neural Network-based Real-time Prediction of {US} Recessions",
  journal = "arXiv preprint arXiv:2310.17571",
  year = "2023"
}

@inproceedings{gibbs2021adaptive,
  author = "Isaac Gibbs and Emmanuel Cand{\`e}s",
  title = "Adaptive Conformal Inference Under Distribution Shift",
  booktitle = "Advances in Neural Information Processing Systems (NeurIPS)",
  volume = "34",
  pages = "1660--1672",
  year = "2021"
}

@inproceedings{tibshirani2019conformal,
  author = "Ryan J. Tibshirani and Rina Foygel Barber and Emmanuel J. Cand{\`e}s and Aaditya Ramdas",
  title = "Conformal Prediction Under Covariate Shift",
  booktitle = "Advances in Neural Information Processing Systems (NeurIPS)",
  volume = "32",
  pages = "2526--2536",
  year = "2019"
}

@article{barber2023beyond,
  author = "Rina Foygel Barber and Emmanuel J. Cand{\`e}s and Aaditya Ramdas and Ryan J. Tibshirani",
  title = "Conformal Prediction Beyond Exchangeability",
  journal = "Annals of Statistics",
  volume = "51",
  number = "2",
  pages = "816--845",
  year = "2023"
}

@article{barber2025predictive,
  author = "Rina Foygel Barber and Ashwin Pananjady",
  title = "Predictive Inference for Time Series: Why is Split Conformal Effective Despite Temporal Dependence?",
  journal = "arXiv preprint arXiv:2510.02471",
  year = "2025"
}

@misc{fred_recprousm156n,
  author = "{Federal Reserve Bank of St. Louis}",
  title = "Smoothed {U.S.} Recession Probabilities [{RECPROUSM156N}]",
  howpublished = "FRED, Federal Reserve Bank of St. Louis",
  note = "\url{https://fred.stlouisfed.org/series/RECPROUSM156N}",
  year = "2026"
}

@techreport{vovk2003mondrian,
  author = "Vladimir Vovk and David Lindsay and Ilia Nouretdinov and Alex Gammerman",
  title = "Mondrian Confidence Machine",
  institution = "On-line Compression Modelling Project (New Series)",
  type = "Technical Report",
  year = "2003"
}

@article{pasche2026extreme,
  author = "Olivier C. Pasche and Henry Lam and Sebastian Engelke",
  title = "Extreme Conformal Prediction: Reliable Intervals for High-Impact Events",
  journal = "Extremes",
  volume = "29",
  pages = "129--155",
  year = "2026"
}

\end{document}